\documentclass[conference]{IEEEtran}

\usepackage[pdftex]{graphicx}
\usepackage{amsmath}
\usepackage{eqparbox}
\usepackage{times}
\usepackage{epsfig}
\usepackage{graphicx}
\usepackage{amsmath}
\usepackage{amssymb}
\usepackage{graphicx}

\usepackage[T1]{fontenc}
\usepackage{courier}
\usepackage{amssymb}
\usepackage{graphicx,multirow}
\usepackage{makecell}
\usepackage{algorithm,algorithmicx,algpseudocode}
\usepackage{booktabs}       

\usepackage{amsmath}
\usepackage{graphics}
\usepackage{xcolor}
\usepackage{epsfig}
\usepackage[T1]{fontenc}
\usepackage{courier}
\usepackage{amssymb}
\usepackage{graphicx,multirow}
\usepackage{makecell}
\usepackage{algorithm,algorithmicx,algpseudocode}
\usepackage{booktabs}       

\usepackage{amsmath}
\usepackage{graphics}
\usepackage{xcolor}
\usepackage{caption}
\usepackage{subcaption}

\graphicspath{ {img/} }

\begin{document}

\title{Semantic Regularization: Improve Few-shot Image Classification by Reducing Meta Shift}
\author{\authorblockN{\authorrefmark{2}Da Chen, \authorrefmark{5}Yongliang Yang, \authorrefmark{3}Zunlei Feng, \authorrefmark{2}Xiang Wu, \\ \authorrefmark{3}MingLi Song, \authorrefmark{5}Wenbin Li, \authorrefmark{2}Yuan He, \authorrefmark{2}Hui Xue, \authorrefmark{2}Feng Mao} \\
\authorblockA{\authorrefmark{2}Alibaba Group} \authorrefmark{3} Zhejiang University \ \authorrefmark{5}University of Bath \ }

\maketitle

\begin{abstract}
Few-shot image classification requires the classifier to robustly cope with unseen classes even if there are only a few samples for each class. Recent advances benefit from the meta-learning process where episodic tasks are formed to train a model that can adapt to class change.
However, these tasks are independent to each other and existing works mainly rely on limited samples of individual support set in a single meta task. This strategy leads to severe meta shift issues across  multiple tasks,  meaning  the  learned  prototypes  or class descriptors are not stable as each task only involves their own support set.
%
%
%
To avoid this problem, we propose a concise Semantic Regularization Network to learn a common semantic space under the framework of meta-learning. In this space, all class descriptors can be regularized by the learned semantic basis, which can effectively solve the meta shift problem.
%
The key is to train a class encoder and decoder structure that can encode the sample embedding features into the semantic domain with trained semantic basis, and generate a more stable and general class descriptor from the decoder.
We evaluate our work by extensive comparisons with previous methods on three benchmark datasets (\textit{MiniImageNet}, \textit{TieredImageNet}, and CUB). The results show that the semantic regularization module improves performance by 4\%-7\% over the baseline method, and achieves competitive results over the current state-of-the-art models.
\end{abstract}

\section{Introduction}
\label{sec:Intro}






\begin{figure*}[ht]
    \centering
    \includegraphics[width=1\linewidth]{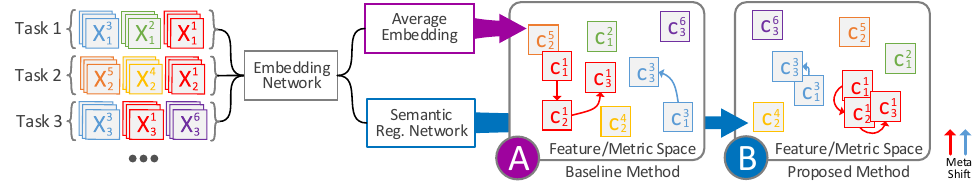}
    \caption{A high-level description of meta shift. $\mathbf{X}_{\mathcal{T}}^{m}$ and $\mathbf{c}_{\mathcal{T}}^{m}$ are the support set and the class descriptor of the $m$-th class in task ${\mathcal{T}}$ respectively. During training, baseline method (\textcolor{magenta}{A}) generates class descriptor that is task-dependent and only concerns the classification result in the current task. The generated class descriptors across tasks are not stable and sparsely distributed (shifting problem), ~{\em e.g.,} the first class is biased due to $\mathbf{c}_{1}^{1}$ and $\mathbf{c}_{2}^{5}$ are very close, query sample in first class may be classified as 5th class. The proposed method (\textcolor{cyan}{B}) learns a common semantic space and regularizes the class descriptor with the learned semantic basis, which solves the meta shift problem effectively. Please refer to Fig.~\ref{fig:cluster} for real examples.
    }
    \label{fig:distance}
\end{figure*}


The human ability to understand new concepts with only a few examples has inspired the research on few-shot learning in recent years.
The main task is to achieve a learnt model that can classify a new category given limited training data. Different from classification models~\cite{he2016residual,simonyan2014vgg,szegedy2017inception_v4} trained on large labelled dataset, few-shot learning model only relies on a few samples of each class (10, 5, or even less). This may easily lead to overfitting during training. To address this issue, Vinyals~\textit{et al.}~\cite{nips2016_vinyals2016matching_net} propose an attention mechanism which can learn an embedding of labelled samples from the support set and achieve good classification performance. This mechanism can be further enhanced by \textit{episodes}, which aim to sub-sample categories and the associated data to simulate few-shot tasks during training.

By investigating transferable embedding from the dataset, as well as the relation between images and the associated category descriptions, existing meta-learning approaches, such as ProtoNet~\cite{snell2017prototypical_net} and RelationNet~\cite{sung2018relation_net}, prove to be effective for few-shot learning. However, they are often restricted by only targeting individual tasks during training, thus cannot efficiently explore the variation of all classes in the whole training set with a comprehensive view. Due to the bias of chosen class subset for an individual training episode, the generated class descriptors/prototypes for the same class in different episodes can be sparsely distributed in the feature space.
As illustrated in Fig.~\ref{fig:distance}-\textcolor{magenta}{A}, the meta task target of ProtoNet~\cite{snell2017prototypical_net},~{\em i.e.,} the class descriptors/prototypes (noted as \textit{descriptor} for the rest of the paper) of different classes, are only required to be distinguishable under certain metric for the current task. However, the descriptors for the same class in different tasks can still be distant to each other in the class domain, as individual tasks are considered separately. In this sense, the class descriptors are not stable among tasks during training, thus the average embedding can easily cause misclassification in the test stage. This is defined as \emph{meta shift} problem.
To solve the meta shift problem, we propose the \textit{Semantic Regularization Network} to learn a common semantic space where the class descriptors can be regularized by the learned semantic basis. The resultant class descriptor can be generally applied to classify various query samples in a stable manner. 
For the input samples, the embedding network firstly extracts the sample embeddings. Then, the class encoder transforms the sample embeddings (class domain) into the semantic representations (semantic domain), which will be regularized by the defined semantic basis. Next, the class decoder transforms the regularized representations back into more stable and general class descriptors (class domain). Finally, The obtained class descriptors are applied to classify the query samples via a metric module which combines a fixed distance metric and a trainable relation module.

We demonstrate that the proposed method achieves state-of-the-art results on three benchmark datasets comparing to classic methods and more recent methods. For 1-shot and 5-shot tasks on \textit{MiniImageNet} and \textit{TieredImageNet}, it improves from 64.12\% to 66.53\%, 80.51\% to 81.66\%, and 66.33\% to 69.82\%, 81.56\% to 85.83\% respectively. For fine-grained dataset CUB, it achieves nearly 5\% improvement on 1-shot task and improves 5-shot result from 81.90\% to 83.53\%. In addition, we illustrate that the proposed semantic regularization module effectively enhances the embedding network to better tackle the covariate shift issue in few-shot learning. Further quantitative and qualitative evaluations show that the proposed method can effectively regularize the descriptor construction and improve classification results.

\section{Related Work}
\label{sec:related_work}


Few-shot learning as an active research topic has been extensively studied.
%
Existing few-shot learning (FSL) methods can be classified into three categories according to~\cite{wang2019generalizing}.
Data-based methods solve FSL problem by augmenting data using prior knowledge.
Model-based methods determine a hypothesis space and constrain the hypothesis space with prior knowledge to approximate the optimal hypothesis.
Algorithm-based methods design the search strategy to find the parameter of the best hypothesis in the hypothesis space with prior knowledge.
In our method, we introduce the Semantic Regularization Network to learn a common semantic space for all classes under the framework of meta-learning, which combines the advantages of model-based and algorithm-based methods.

A number of model-based methods aim to improve the robustness of the training process for few-shot learning.
Garcia~\textit{et al.}~\cite{garcia2017few_graph_gnn} propose a graph neural network according to the generic message-passing inference method. Zhao~\textit{et al.}~\cite{zhao2018msplit_few_shot} split the features to three orthogonal parts to improve the classification performance for few-shot learning, allowing simultaneous feature selection and dense estimation. Chen~\textit{et al.}~\cite{chen2018a} present a comprehensive analysis and investigate the cross-domain generalization ability for many existing few-shot learning methods. They also propose a new method which achieves state-of-the-art result on the CUB dataset~\cite{WahCUB_200_2011}. 
Chen~\textit{et al.}~\cite{chen2019aug_few_shot} propose a Self-Jig algorithm to augment the input data in few-shot learning by synthesizing new images that are either labelled or unlabelled. Chu~\textit{et al.}~\cite{chu_cvpr2019_spot_and_learn} augment the input images by extracting varying sequences of patches on every forward-pass with discriminative observed information using maximum entropy reinforcement learning.

For algorithm-based methods, a popular strategy is meta-learning (also called learning-to-learn) with multi-auxiliary tasks~\cite{douze2018low_large_scale_diffusion,finn2017maml,hariharan2017low_hallucinating_features,jiang2018learning_caml,mishra2017snail,qiao2018few_shot_activations,sung2018relation_net,nips2016_vinyals2016matching_net,wang2018low_imaginary_data}.
The key insight is how to robustly accelerate the network learning progress without overfitting on limited training data. Finn~\textit{et al.}~\cite{finn2017maml} propose MAML to search for the best initial weights through gradient descent for the network training, making the fine-tuning of the network easier. REPTILE~\cite{nichol2018reptile} simplifies the complex computation of MAML by incorporating a $L_2$ loss, but still performs in high dimensional space. To reduce the complexity, Rusu~\textit{et al.}~\cite{rusu2018LEO} propose a network called LEO to learn a low dimension latent embedding of the model. CAML~\cite{jiang2018learning_caml} extends MAML by partitioning the model parameters into context parameters and shared parameters, enabling a larger network without over-fitting.

\begin{figure*}[h]
    \centering
    \includegraphics[width=0.95\linewidth]{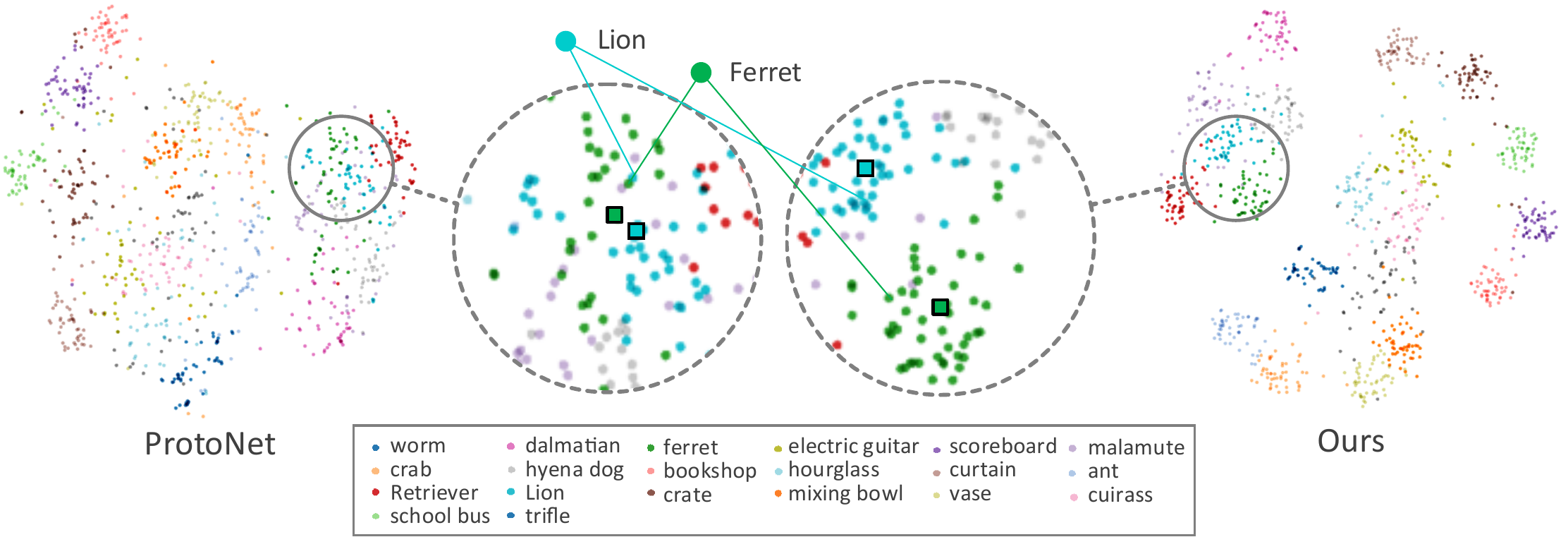}
    \caption{t-SNE plot of latent space features for 20 classes in test set in \textit{MiniImageNet}. 50 samples are randomly selected from each classes. Feature point color corresponds to class label in the bottom box. Squares denote example class descriptors.}
    \label{fig:cluster}
\end{figure*}

Another stream of meta-learning based approaches~\cite{oreshkin2018tadam,snell2017prototypical_net,sung2018relation_net,nips2016_vinyals2016matching_net} attempt to learn a deep embedding model that can effectively project the input samples to a specific feature space. Then the samples can be classified by the nearest neighbour criterion using a distance function such as Cosine distance or Euclidean distance,~{\em etc.}
Koch~\textit{et al.}~\cite{koch2015siamese_few_shot} propose the Siamese network to extract embedding features from input images and converge images in the same class.
Matching Network~\cite{nips2016_vinyals2016matching_net} utilizes an augmented neural network for feature embedding, forming the basis for metric learning.
The most relevant work to ours is ProtoNet~\cite{snell2017prototypical_net}. It proposes to model the class descriptor of each class with a simple average pooling on embedding sample features. However, the distance estimation only concerns the current task but not the whole train set. Many approaches extend ProtoNet by improving its class descriptor. RelationNet~\cite{sung2018relation_net} exploits a relation network with a learnable non-linear comparator instead of a fixed linear one. Lee~\textit{et al.}~\cite{lee2019metaOpt} propose to use discriminatively trained linear predictors as learners to learn a more generalized embeddings under a linear classification rule. Li~\textit{et al.}~\cite{li2019metaCTM} focus on all the class samples in the support set of each task and present a better feature for metric learning.  TADAM~\cite{oreshkin2018tadam} produces a task-dependent metric space based on conditioning a learner on the task set. Li~\textit{et al.}~\cite{li_cvpr2019_revisiting} propose to replace the distance measured at the image level to a local descriptor based image-to-class measure. Liu~\textit{et al.}~\cite{liu2018learning_TPN} propose a transductive propagation network (TPN) which learns both the parameters of feature embedding and the graph construction. Unlike methods mentioned above which fine-tune the embedding network directly, Sun~\textit{et al.}~\cite{sun2019meta_mtl} propose to apply a neuron-level scaling and shifting on the embedding network with a hard-task meta batch setting. 

In this paper, we propose a novel semantic regularization network under the meta-learning framework. This network is able to learn a common semantic space where all classes can be regularized by the semantic basis in that space. This avoids the meta shift problem and achieves state-of-the-art performance.



\section{Semantic Regularization Network}
\label{sec:method}

In this section, we first present preliminaries in the meta-learning setting with independent episodic tasks and the subsequent meta shift problem, then give an overview of our approach, finally elaborate the details of our method.

\subsection{Meta-learning for few-shot learning}
\label{subsec:meta_learing_general}

Few-shot learning is a challenging problem as it is required to classify unseen queries with only limited data for training.
One solution is to apply a meta-learning process composed by multiple $M$-way $K$-shot episodic classification tasks~\cite{nips2016_vinyals2016matching_net}. For each classification task, we have $M$ classes with $K$ samples from each class. Formally, given a specific task $\mathcal{T}$, a support set $\mathcal{S}$ and a query set $\mathcal{Q}$ are randomly selected from the entire training dataset:~\textbf{a}) the support set is denoted by $\mathcal{S}=\{\mathbf{X}^1,\mathbf{X}^2,\dots,\mathbf{X}^m,\dots,\mathbf{X}^M\}$, where each class set $\mathbf{X}^m=\{x^m_1,x^m_2,\dots,x^m_K\}$ contains $K$ samples, and $l^m$ is the corresponding label of class $\mathbf{X}^m$; \textbf{b}) the query set is $\mathcal{Q}=\{x^m_q\}$.
The following method is under this meta-learning setting.

\subsection{Meta shift problem}

During the meta-learning process, most of the existing methods only focus on the selected samples in the support and query sets ($\mathcal{S}$ and $\mathcal{Q}$) of the current task $\mathcal{T}$ ({\em e.g.,} 5-way, 5-shot classification), while ignoring the overall sample distribution within the entire training set.
%
%
We spot that for different training tasks, support samples from the same class can be different as they are picked randomly from the entire training set. This leads to unstable episodic task training in meta-learning. This unstable issue among meta-learning tasks is defined as meta shift.

For example, as mentioned in Section~\ref{sec:related_work}, the recent popular framework ProtoNet utilizes a feature embedding network to map the $m$-th class samples $\mathbf{X}^m=\{x^m_k ,k\in[1,2,\dots,K]\}$ to a mean vector $c^m$. They take this as a class descriptor in the common feature space (named class domain in this paper):
\begin{equation}
\label{equ:pnet_ck}
    c^m = \frac{1}{K}\sum^K_{k=1} f_{\mathrm{EM}} (x^m_k),
\end{equation}
where $f_{\mathrm{EM}}$ is the embedding function.

As shown in Fig.~\ref{fig:cluster}-Left, sample embeddings in the same class from ProtoNet are sparsely distributed. In a $K$-shot task, $K$ sample embeddings are randomly chosen from this feature space and averaged as a class descriptor. It is clear that a query sample from similar class may be mis-classified (lion vs. ferret, ant vs. worm,~\textit{etc}.). This is because ProtoNet only focuses on the current task during training, thus the generated class descriptors are sparsely distributed in the class domain and are only constrained to fit the current task. This can easily lead to meta-shift problem (see Fig.~\ref{fig:distance}-\textcolor{magenta}{A}).

%
%
Different from the above, we aim at a better regularization over all the training tasks using a Semantic Regularization Network. This network is supposed to resolve the meta shift problem and generate more stable class descriptors (see Fig.~\ref{fig:distance}-\textcolor{cyan}{B}) that are distinctive in the feature space (see Fig.~\ref{fig:cluster}-Right), which in turn results in significantly improved performance for few-shot classification.

\subsection{Key idea}
\label{subsec:key_idea}


%
%

To solve the meta shift problem, we consider how humans classify objects with few examples. Conceptually, humans can find the key components from objects and perform classification accordingly. 
For instance, the feather of wings, the shape of neck, mouth, and tail, \textit{etc.}, can be very helpful to classify birds. These key components can be treated as semantic basis to form a semantic domain, and all classes can be regularized in this domain according to the basis.

\begin{figure*}[h]
    \centering
    \includegraphics[width=1\linewidth]{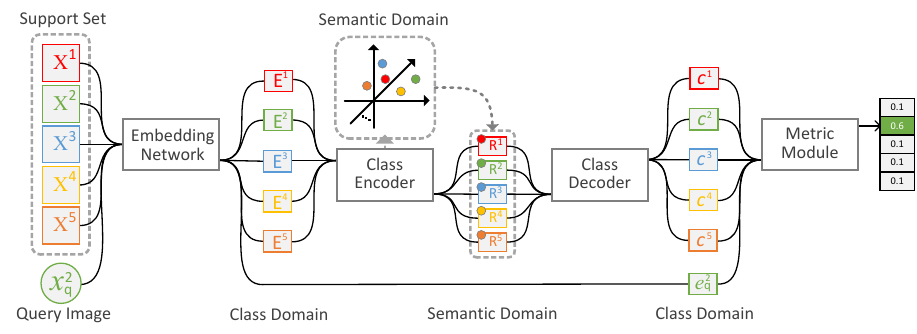}
    \caption{Overview of the proposed Semantic Regularization Network. The Network is composed by four major parts: A feature \textbf{Embedding Network} ($f_{\mathrm{EM}}$), \textbf{Class Encoder} ($f_{\mathrm{EN}}$) for regularizing class representation with semantic basis in semantic domain, \textbf{Class Decoder} ($f_{\mathrm{DE}}$) to transform the regularized representation back to class domain and form a stable and general class descriptor and \textbf{Metric Module} to compare class descriptor and query sample embedding.}
    \label{fig:net}
\end{figure*}

With the above inspiration, we propose to use a class encoder-decoder structure to realize semantic regularization (see Fig.~\ref{fig:net}-Middle). The encoder forms a semantic domain with trained semantic basis, and then aggregate the sample features (within the same class) from the class domain to the semantic domain. As there are much more samples of each class in the whole training set compared with individual support set, the embedding features are redundant to comprehensively describe a class. During the training process with multiple tasks, the aggregated sample features keep updating the encoder on how to form a better class representation in the semantic domain with semantic basis. Such representation can effectively avoid task-dependent feature causing the meta shift problem, and describe various classes in a more stable way as shown by the cyan curve in Fig.~\ref{fig:test}-a. Each class representation in the semantic domain is further reconstructed back into the class domain to form a more stable and general class descriptor to classify query samples. Benefiting from a robust embedding network and a stable class descriptor, the proposed method largely improves classification performance. 

\subsection{Proposed method}
\label{subsec:adnet}

In this subsection, we present the details of the Semantic Regularization Network, which contains the embedding network, the class encoder-decoder network, and the metric module. As shown in Fig.~\ref{fig:net},
the embedding network first extracts sample embeddings from image samples in the support/query set. Then the class encoder transforms the sample embeddings (in class domain) into the semantic representations (in semantic domain), which will be regularized by the learned semantic basis. The class decoder transforms the regularized representations (denoted by `activations' ) back into  more stable and general class descriptors (in class domain). Finally, the metric module is adopted to measure the similarity between the embeddings of query sample and the class descriptors.

\subsubsection{Embedding samples.} Similar to previous metric learning based methods, the input samples are first fed into an embedding network.

Under the meta-learning framework,
given the $m$-th class samples $\mathbf{X}^m=\{x^m_1, x^m_2, ..., x^m_K\}$ ($K$-shot task) of the support set $\mathcal{S}=\{\mathbf{X}^1, \mathbf{X}^1,\dots, \mathbf{X}^m, \dots, \mathbf{X}^M\}$, the embedding network $f_{\mathrm{EM}}$ extracts $K$ embeddings $\mathbf{E}^m=\{\mathbf{e}^m_1, \mathbf{e}^m_2, ..., \mathbf{e}^m_K\}$ defined as follows:
\begin{equation}
\label{equ:embedding}
\begin{split}
    \mathbf{e}^m_k &=  f_{\mathrm{EM}} (x^m_k), \  x^m_k\in \mathbf{X}^m, \  k\in [1,2,\dots,K],
\end{split}
\end{equation}
where $\mathbf{e}^m_k$ is the extracted embedding of the input sample $x^m_k$.

\subsubsection{Encoding classes.}
\label{subsubsec:encoding}

As described above, humans can find key components of the objects and perform classification accordingly. These key components can be treated as semantic basis to form a semantic domain, and all classes can be regularized in the semantic domain. So, we define $N$ semantic basis $\mathbf{B}=\{\mathbf{b}_1, \mathbf{b}_2, ..., \mathbf{b}_N\}$, where each base vector $\mathbf{b}_i$ is $D$-dimensional. The semantic basis is optimizable in the training stage.
To regularize the class embeddings $\mathbf{E}^m$, we devise a class encoder $f_{\mathrm{EN}}$, which can transforms class embeddings into semantic domain as follows:
\begin{equation}
\label{equ:encoder1}
\begin{split}
    \mathbf{r}^m_k = f_{\mathrm{EN}}(\mathbf{e}^m_k), \mathbf{e}^m_k \in \mathbf{E}^m.
\end{split}
\end{equation}
With Eqn. (\ref{equ:encoder1}), we can get $K$ transformed representations $\mathbf{R}^m=\{\mathbf{r}^m_k,k\in[1,2,\dots,K]\}$ of the $m$-th class.
Then, given the representations $\mathbf{R}^m$ and the semantic basis $\mathbf{B}$, the activations $\mathbf{A}^m=\{\mathbf{a}^m_1,\mathbf{a}^m_2,...,\mathbf{a}^m_i,...,\mathbf{a}^m_N\}$ between the representations and semantic basis can be defined as follows:
\begin{equation}
\label{equ:activate}
\begin{split}
\mathbf{a}^m_i = \frac{1}{K}\sum_{k=1}^{K}v_i(\mathbf{r}^m_k)(\mathbf{r}^m_k - \mathbf{b}_i),
\end{split}
\end{equation}
where the residual $(\mathbf{r}^m_k - \mathbf{b}_i)$ denotes the activation between the $k$-th transformed representation $\mathbf{r}^m_k$ and the $i$-th base vector $\mathbf{b}_i$, $\mathbf{a}^m_i$ is the average activation between the $m$-th class representations and the $i$-th base vector,  $v_i(\mathbf{r}^m_k)$ denotes the activation's intensity between the representation $\mathbf{r}^m_k$ and the base vector $\mathbf{b}_i$. Inspired by~\cite{arandjelovic2016netvlad},  $v_i(\mathbf{r}^m_k)$ should has the ability to express the relative intensity among all the semantic basis. Thus $v_i(\mathbf{r}^m_k)$ is defined as follows:
\begin{equation}
\label{equ:intensity}
v_i(\mathbf{r}^m_k) = \frac{e^{-\alpha \left \| \mathbf{r}^m_k - \mathbf{b}_i\right \|^2}}{\sum^N _{i=1} e^{-\alpha \left \| \mathbf{r}^m_k - \mathbf{b}_i\right \|^2}},
\end{equation}
where parameter $\alpha$ controls the sensitivity to the residual.

It can be seen that for all classes in the support set $\mathcal{S}$,  the class representations $\mathbf{R}^m$ can be expressed by accumulating the representations' activations on each semantic basis in the semantic domain.

\subsubsection{Decoding classes.}

With the activations $\mathbf{A}^m$ of the $m$-th class representations, we devise a class decoder to transform the activations into the $m$-th class descriptor from semantic domain to class domain. The transformed class descriptor is expected to be more general and stable. In the class decoder, the transformation function $f_{\mathrm{DE}}$ generates the $m$-th class descriptor $\mathbf{c}^m$ with the activations $\mathbf{A}^m$ as follows:
\begin{equation}
\label{equ:decoder}
\mathbf{c}^m = f_{\mathrm{DE}} (\mathbf{A}^m),
\end{equation}
where $\mathbf{c}^m$ share a same size with query embedding feature $e_q$.

In summary, with the above class encoder-decoder, the $K$ embeddings $\mathbf{E}^m$ of the $m$-th class are transformed into a general and stable class descriptor $\mathbf{c}^m$. With the class encoder, the class embeddings $\mathbf{E}^m$ are firstly transformed into the semantic representations $\mathbf{R}^m$. Then, the activations of the semantic representation $\mathbf{R}^m$ on the semantic basis $\mathbf{B}$ are calculated with Eqn. (\ref{equ:activate}), which expresses the $K$ samples using related semantic basis. Finally, the activations $\mathbf{A}^m$ are transformed back into the class domain, which generates a general and stable class descriptor $\mathbf{c}^m$.

\subsubsection{Metric module.}

With above embedding network, class encoder and class decoder, the support set $\mathcal{S}=\{\mathbf{X}^1,\mathbf{X}^2,\dots,\mathbf{X}^m,\dots,\mathbf{X}^M\}$ can be transformed into $K$ class descriptors $\{\mathbf{c}^m, m \in [1,2,\dots,M]\}$. Meanwhile, the query image $\mathbf{x}^m_q$ of query set $\mathcal{Q}$ is extracted as embedding $\mathbf{e}^m_q$ with the embedding network.

It remains crucial on how to measure the similarity between the query embedding $\mathbf{e}^m_q$ and the $K$ class descriptors $\{\mathbf{c}^m, m \in [1,2,\dots,M]\}$. Existing works, such as ProtoNet~\cite{snell2017prototypical_net}, usually adopt a simple fixed distance metric,~{\em i.e.,} Cosine distance, Euclidean distance,~{\em etc.} In this work, we utilize a metric module which includes a general relation module~\cite{sung2018relation_net} along with a fixed distance metric via a co-training mechanism, resulting in a more robust similarity measurement for comparing the query embedding $\mathbf{e}^m_q$ to the general class descriptors $\{\mathbf{c}^m, m \in [1,2,\dots,M]\}$. Here we discuss the metric module in details.


Similar to ProtoNet~\cite{snell2017prototypical_net}, we employ an Euclidean distance function and produce a distribution over all classes given a query image $\mathbf{x}^m_q$ from the query set $\mathcal{Q}$. The distribution is based on a softmax over the distance between the query embedding $\mathbf{e}^m_q$ and the general class descriptors $\{\mathbf{c}^m, m \in [1,2,\dots,M]\}$. The loss function can be defined as:
\begin{equation}
\label{equ:loss2}
\begin{split}
 \mathcal{L}_{1} = \mathbb{D}(\mathbf{e}^m_q,\mathbf{c}^m) + \log\sum^{M}_{m'=1}\exp[-\mathbb{D}(\mathbf{e}^m_q,\mathbf{c}^{m'})],\  m'\neq m,
\end{split}
\end{equation}
where $\mathbb{D}$ denotes the Euclidean distance metric, and $\mathbf{e}^m_q = f_{\mathrm{EM}}(\mathbf{x}^m_q)$.

Apart from the fixed distance function, we also include a trainable relation module. Given the query embedding $\mathbf{e}^m_q$ and class descriptor $\mathbf{c}^m$, the output of the relation module indicates the correlation between them. We treat this output as the relation score~\cite{sung2018relation_net} among $M$ classes and all query images.
Similar to Eqn. (\ref{equ:loss2}), given the correct class label $l^m$ for $\mathbf{x}^m_q$, the loss for the relation module can be written as:

\begin{equation}
\label{equ:relation_mse_loss3}
\begin{split}
\mathcal{L}_{2} &= \mathbb{E}[f_{RS}(\mathbf{e}^m_q\oplus\mathbf{c}^m),l^m] \\ &+ \log\sum^M_{m'=1}\exp\{-\mathbb{E}[f_{RS}(\mathbf{e}^m_q\oplus\mathbf{c}^{m'}),l^{m}]\},\  m'\neq m,
\end{split}
\end{equation}
where $f_{RS}$ is the relation function, $\oplus$ denotes concatenation, $\mathbb{E}$ indicates the Cross Entropy metric.

The total loss $\mathcal{L}$ of the Semantic Regularization Network can then be summarized as follows:
\begin{equation}
\label{equ:total_loss}
\mathcal{L} = \alpha * L_{1} + \beta * L_{2} + \gamma *\left \| \Theta \right \|^2,
\end{equation}
where $\alpha, \beta, \gamma$ are the weights of the loss $L_{1}$, $L_{2}$ and the regularization term, $\Theta = [\theta_{\mathrm{EN}}, \theta_{\mathrm{DE}}, \theta_{\mathrm{RM}}]$ are the training parameters of class encoder, class decoder and relation module. The algorithmic process of this module and the entire network can be found in the supplementary material.

\section{Evaluation}
\label{sec:exper_result}

In this section, we first describe the network architecture and the datasets in our evaluation, then detail the training procedure and a thorough discussion of the advantages over the most relevant work~\cite{snell2017prototypical_net}.

\subsection{Network architecture}
\label{subsec:archi_ADNet}


The overall architecture of the proposed network is shown in Fig.~\ref{fig:net}. Noted that it can incorporate different feature embedding networks. In our experiments, we apply three embedding networks that are usually used in the few-shot image classification field~\cite{li_cvpr2019_revisiting,sun2019meta_mtl}, including 4 Conv with four convolutional blocks, ResNet12~\cite{he2016residual} with three residual blocks followed by an average pooling layer and WRN with layer depth 28  and a widening factor of 10~\cite{zagoruyko2016wide_residul_WRN}. The class encoder has a convolutional block and a softmax operator for soft-assignment (see details in the supplemental) to produce a sparse matrix which includes all weights for semantic basis. With a multiplication between the weights and the residuals followed by a normalization layer, the class representation composed by semantic basis in the semantic domain is obtained. The decoder exploits a normal convolutional block to transform the class representation from the semantic domain to general class descriptor in the class domain. To compare the obtained general class descriptor with the embedding features of the samples in the query set, a fixed distance metric ({\em i.e.,} Euclidean distance) and a learnable relation module (two Conv + two FC) based similarity measurement are jointly applied, while a cross-entropy loss is employed as classification loss. The network then updates the metric module along with the encoder-decoder structure, and fine-tune the feature embedding network simultaneously.

\subsection{Dataset}
\label{subsec:dataset}

\textit{MiniImageNet} dataset, as proposed in~\cite{nips2016_vinyals2016matching_net}, is a benchmark to evaluate few-shot learning methods. This dataset is a subset from \textit{ImageNet}. \textit{MiniImageNet} contains 60,000 images from only 100 classes, and each class has 600 images. We also follow the data split strategy in~\cite{ravi2016optimization} to sample images of 64 classes for training, 16 classes for validation, 20 classes for test.

\textit{TieredImageNet}~\cite{ren2018meta_semi_tiered} is also a subset from \textit{ImageNet} but with more classes. It contains 779,165 images with 608 classes, and is split into (351, 97, 160) classes as training, validation and test sets. Different from \textit{MiniImageNet}, the split is based on WordNet~\cite{miller1995wordnet} to ensure the above sets are semantically unrelated.

\textit{Caltech-UCSD CUB-200-2011} dataset~\cite{WahCUB_200_2011} is a dataset for fine-grained classification. The CUB-200-2011 dataset contains 200 classes of birds with 11,788 images in total. For few-shot learning classification task, we follow the split in~\cite{hilliard2018few_conditional_embedding_maco} for evaluation. 200 species of birds are split to 100 classes for training, 50 classes for validation, and 50 classes for test.

\begin{table*}[t]
\centering
\scalebox{1}{
\begin{tabular}{l|c|c c|c c}
\Xhline{3\arrayrulewidth}
\multicolumn{2}{c|}{ }  & \multicolumn{2}{c|}{\textit{MiniImageNet}} & \multicolumn{2}{c}{\textit{TieredImageNet}} \\
\Xhline{2\arrayrulewidth}
\textbf{Model}  & \textbf{Embed. Net.} & \textbf{1-Shot} & \textbf{5-Shot} & \textbf{1-Shot} & \textbf{5-Shot}\\ 
\Xhline{2\arrayrulewidth}
MatchingNet~\cite{nips2016_vinyals2016matching_net}  & 4 Conv   & 43.56 $\pm$ 0.84\%   & 55.31 $\pm$ 0.73\%   & -    &    -      \\ 
MAML~\cite{finn2017maml}      & 4 Conv            & 48.70 $\pm$ 1.84\%   & 63.11 $\pm$ 0.92\%   &   51.67 $\pm$ 1.81\%      &   70.30 $\pm$ 1.75\%     \\ 
RelationNet~\cite{sung2018relation_net} & 4 Conv           & 50.44 $\pm$ 0.82\%   & 65.32 $\pm$ 0.70\%   & 54.48 $\pm$ 0.93\%   & 71.32 $\pm$ 0.78\%          \\ 
REPTILE~\cite{nichol2018reptile}      &   4 Conv       & 49.97 $\pm$ 0.32\%   & 65.99 $\pm$ 0.58\%     & 48.97 $\pm$ 0.21\% & 66.47 $\pm$ 0.21\%           \\ 
ProtoNet~\cite{snell2017prototypical_net} &     4 Conv         & 49.42 $\pm$ 0.78\%   & 68.20 $\pm$ 0.66\%   & 53.31 $\pm$ 0.89\% & 72.69 $\pm$ 0.74\%     \\ 
Baseline*~\cite{chen2018a}  &       4 Conv        & 41.08 $\pm$ 0.70\%         & 54.50\% $\pm$ 0.66         & -    & -      \\ 
Spot\&learn~\cite{chu_cvpr2019_spot_and_learn}  &       4 Conv        & 51.03 $\pm$ 0.78\%         & 67.96\% $\pm$ 0.71       & -   & -       \\ 
DN4~\cite{li_cvpr2019_revisiting}  &       4 Conv        & 51.24 $\pm$ 0.74\%         & 71.02\% $\pm$ 0.64\%    & -    & -      \\ 

Discr. k-shot~\cite{bauer2017discriminative_k_shot} & ResNet34  & 56.30 $\pm$ 0.40\%   & 73.90 $\pm$ 0.30\%   & -  & -  \\ 
Self-Jig~\cite{chen2019aug_few_shot}     &   ResNet50
& 58.80 $\pm$ 1.36\%   & 76.71 $\pm$ 0.72\%   & -   & -       \\ 
SNAIL~\cite{mishra2017snail}    & ResNet12
& 55.71 $\pm$ 0.99\%   & 68.88 $\pm$ 0.92\%   & - & -            \\ 


CAML~\cite{jiang2018learning_caml}      &  ResNet12      & 59.23 $\pm$ 0.99\%   & 72.35 $\pm$ 0.71\%   & -   & -       \\ 

TPN~\cite{liu2018learning_TPN}        &   ResNet12
& 59.46\%         & 75.65\%  & -       & -   \\ 
MTL~\cite{sun2019meta_mtl}        & ResNet12
& 61.20 $\pm$ 1.8\%   & 75.50 $\pm$ 0.8\%   & -             \\
DN4~\cite{li_cvpr2019_revisiting}        & ResNet12
& 54.37 $\pm$ 0.36\%   & 74.44 $\pm$ 0.29\%   & -   & -          \\

TADAM~\cite{oreshkin2018tadam}      &    ResNet12
& 58.50\%         & 76.70\%         & -  & -        \\

MetaOpt~\cite{lee2019metaOpt}        & ResNet12
& 62.64 $\pm$ 0.61\%   & 78.63 $\pm$ 0.46\%   & 65.99 $\pm$ 0.72\%  & 81.56 $\pm$ 0.53\%          \\
CTM~\cite{li2019metaCTM}        & ResNet18
& 64.12 $\pm$ 0.82\%   & 80.51 $\pm$ 0.13\%   & 64.78 $\pm$ 0.11\%   & 81.05 $\pm$ 0.52\%       \\
Qiao~\cite{qiao2018few_shot_activations}   &    WRN28
& 59.60 $\pm$ 0.41\%   & 73.74 $\pm$ 0.19\%   & -    & -      \\ 
LEO~\cite{rusu2018LEO}        & WRN28
& 61.76 $\pm$ 0.08\%   & 77.59 $\pm$ 0.12\%   & 66.33 $\pm$ 0.05\%    & 81.44 $\pm$ 0.09\%        \\
\Xhline{3\arrayrulewidth}
\textbf{Ours\_Res}      &  ResNet12   & 63.37 $\pm$ 0.14\%  & 79.03 $\pm$ 0.16\%   & 65.83 $\pm$ 0.13\%  & 82.26 $\pm$ 0.22\% \\
\textbf{Ours\_WRN}      &  WRN28  & \textbf{66.53 $\pm$ 0.15\%}  & \textbf{81.66 $\pm$ 0.18\%}   & \textbf{69.82 $\pm$ 0.22\%}  & \textbf{85.83 $\pm$ 0.13\%} \\
\Xhline{3\arrayrulewidth}
\end{tabular}
}
\caption{Few-shot classification accuracy results on \textit{MiniImageNet} and \textit{TieredImageNet} on 5-way 1-shot, 5-way 5-shot tasks. All accuracy results are reported with 95\% confidence intervals.
For each task, the best-performing method is highlighted. `-': the results are not reported. }
\label{tab:Mini_result}
\end{table*}

\begin{table*}[]
\centering
\scalebox{1}{
\begin{tabular}{l|c c c}
\Xhline{3\arrayrulewidth}
\textbf{Model}      & Embed. Net. & \textbf{5-Way 1-Shot} & \textbf{5-Way 5-Shot} \\ 
\Xhline{2\arrayrulewidth}
MatchingNet~\cite{nips2016_vinyals2016matching_net} & 4 Conv & 61.16 $\pm$ 0.89   & 72.86 $\pm$ 0.70   \\ 
MAML~\cite{finn2017maml}    &  4 Conv  & 55.92 $\pm$ 0.95\%   & 72.09 $\pm$ 0.76\%   \\ 
ProtoNet~\cite{snell2017prototypical_net} & 4 Conv  & 51.31 $\pm$ 0.91\%   & 70.77 $\pm$ 0.69\%   \\ 
MACO~\cite{hilliard2018few_conditional_embedding_maco}   &  4 Conv   & 60.76\%        & 74.96\%        \\ 
RelationNet$^{+}$~\cite{sung2018relation_net} & 4 Conv & 62.45 $\pm$ 0.98\%   & 76.11 $\pm$ 0.69\%   \\ 
ProtoNet$^{+}$~\cite{snell2017prototypical_net} & 4 Conv & 63.52 $\pm$ 0.25\%   & 79.06 $\pm$ 0.20\%   \\ 
Baseline++~\cite{chen2018a} & 4 Conv & 60.53 $\pm$ 0.83\%   & 79.34 $\pm$ 0.61\%   \\ 
DN4-DA~\cite{li_cvpr2019_revisiting} & 4 Conv & 53.15 $\pm$ 0.84\%   & 81.90 $\pm$ 0.60\%   \\ 
\Xhline{3\arrayrulewidth}
\textbf{Ours\_conv}    & 4 Conv   & \textbf{67.85  $\pm$ 0.24\%}  & \textbf{83.53 $\pm$ 0.16\%}   \\
\Xhline{3\arrayrulewidth}
\end{tabular}
}
\caption{Few-shot classification accuracy results on CUB dataset~\cite{WahCUB_200_2011} on 5-way 1-shot task, 5-way task 5-shot. All accuracy results are reported with 95\% confidence intervals.
For each task, the best-performing method is highlighted. `+': re-implementation results for a fair comparison.}
\label{tab:CUB_result}
\end{table*}

\subsection{Training details}
\label{subsec:training_details}

Several recent works show that a typical training process can include a pre-trained network~\cite{qiao2018few_shot_activations,rusu2018LEO} or employ co-training~\cite{oreshkin2018tadam} for feature embedding. This can significantly improve the classification accuracy. In this paper, we adapt the training strategy from~\cite{qiao2018few_shot_activations} to pre-train the feature embedding network. In our training process, we follow the standard setup as applied by most few-shot learning frameworks, introduced in Section~\ref{subsec:meta_learing_general}, to train the embedding network, class encoder and decoder along with the metric module. Stochastic Gradient Descent (SGD) is used as the optimizer for ResNet12 and WRN, while ADAM is chosen as the optimizer for 4 Conv. During testing, we follow the setup in~\cite{snell2017prototypical_net}, 15 query images per class are batched in the testing episode. The accuracy is obtained by averaging 600 test tasks which generate test batches with random classes and random samples. The training parameters with different embedding networks on all datasets are presented in the supplementary material. The training time for the proposed method is close to ProtoNet (both under ResNet12) as the semantic regularization module has a simple architecture (~75 seconds per epoch on a single P100 machine for both methods). The number of semantic basis is an important hyper-parameter. See supplemental for its relation with classification accuracy.

\subsection{Results}
\label{subsec:eval_result}

As detailed in Table~\ref{tab:Mini_result}, the proposed method is compared with models including classic methods and recent methods under two tasks (5-way 1-shot, 5-way 5-shot) on two datasets (\textit{MiniImageNet} and \textit{TieredImageNet}). It can be seen that our method achieves state-of-the-art results. 
%
%
In particular, our method yields a margin of increment over the current state-of-the-art approach LEO~\cite{rusu2018LEO} under WRN embedding network in all tasks on two datasets. For example, in \textit{TieredImageNet}, the results in two tasks from our method and LEO are 69.82\% vs. 66.33\% and 85.83\% vs. 81.44\%. Furthermore, the proposed method achieve competitive result comparing to MetaOpt~\cite{lee2019metaOpt} and CTM~\cite{li2019metaCTM} which are the state-of-the-art results with ResNet. Our result with ResNet in \textit{MiniImageNet} is slightly worse than the result of CTM as their embedding network is deeper. On the other hand, our result is better in tasks on \textit{TieredImageNet} even with less complex embedding network.
%
%
Note that some of the methods even apply a deeper feature embedding network with much more parameters to train, such as ResNet-50 used by Sel-Jig~\cite{chen2019aug_few_shot}, ResNet-34 used by Discriminative k-shot~\cite{bauer2017discriminative_k_shot}. Even so, our method still achieves better performance applying a simpler embedding network.

Table~\ref{tab:CUB_result} summarizes the comparisons on the CUB dataset. Our method yields the highest accuracy from all trials. In the 5-way 1-shot test, we achieves $67.85\%$ accuracy which improves to the classic MatchingNet~\cite{nips2016_vinyals2016matching_net} by $6.69\%$. More significant improvement is achieved for the 5-shot 5-way test. Our classification accuracy is $83.53\%$, an improvement of $4.47\%$ over ProtoNet$^{+}$. Comparing to the recently proposed Baseline++~\cite{chen2018a}, our method demonstrates a significant improvement of $7.32\%$ and $4.19\%$ in tests.


\subsection{Effectiveness of semantic regularization}
\label{subsec:ablation}



\begin{figure*}[t]
\centering
\begin{subfigure}{.47\textwidth}
  \centering
  \includegraphics[width=\linewidth]{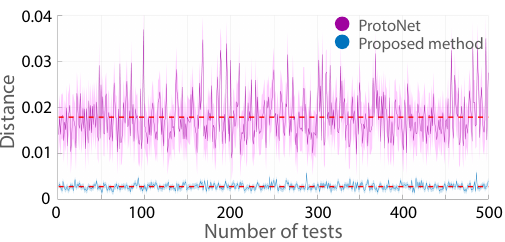}
  \caption{Meta shift measurement}
  \label{fig:curve_fig}
\end{subfigure}%
\begin{subfigure}{.53\textwidth}
  \centering
  \includegraphics[width=0.90\linewidth]{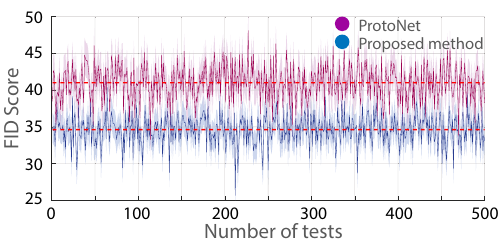}
  \caption{FID score}
  \label{fig:FID}
\end{subfigure}
\caption{Effectiveness of the proposed Semantic Regularization Network compared with baseline model ProtoNet. (a) Meta shift measurement: The statistics on the distance between class descriptors of class `worm' and their mean in 500 tests for baseline method and the proposed method. (b) Fr\'{e}chet Inception Distance (FID) score: In each of the 500 tests as above, FID score measures the similarity between training and testing feature distribution for the few-shot task. Details are presented in Section~\ref{subsec:ablation}.}
\label{fig:test}
\end{figure*}

\textbf{Meta shift.} To highlight the advantage of the proposed method against baseline model~{\em i.e.,} ProtoNet~\cite{snell2017prototypical_net}, we measure the stability of the class descriptors of the proposed method against other methods having meta shift issue. Note that for a fair comparison, we use adapted ProtoNet$^+$ with the same pre-trained embedding network. For each class in the test set of \textit{MiniImageNet}, we randomly pick 5 samples to construct the class descriptor. We repeat this process for 500 tests to generate 500 class descriptors for each baseline. The Euclidean distance from the `mean' class descriptor to a specific class descriptor in each test indicates the variation of the generated class descriptors. Fig.~\ref{fig:test}-a shows statistics for the class `worm'. The proposed method (cyan curve with much less variation) is able to generate more stable class descriptors comparing to ProtoNet$^+$ (magenta curve with severe feature shifting). This results in better performance for the few-shot classification task. More results of the effectiveness of semantic regularization and comparisons of meta shift express in the supplementary material.

\noindent\textbf{Covariate shift.} Given limited training data (even one sample per class for 1-shot task), when dealing with unseen classes during meta testing, the training data distribution is likely to be different from the distribution of the query data (covariate shift~\cite{sugiyama2007covariate_shift_weighted_cross_val}). It requires a generalized embedding network to embed them into a similar feature distribution and be classified with a stable class descriptor from training.

Similar to the qualitative evaluation in~\cite{rusu2018LEO}, we use t-SNE projection to visualize the feature space of ProtoNet$^+$ and our method.
For clarity, we randomly choose 50 samples from each class in the test set and extract their features from the embedding network. The t-SNE plot of the feature space is shown in Fig.~\ref{fig:cluster} (the plot of all samples and the plot for CUB dataset are included in the supplementary material).
%
The feature space of the proposed method is observed to be different from the one of ProtoNet$^+$ in both \textit{MiniImageNet} and CUB dataset. Our embedding features are densely clustered within each class while sparsely distributed among different classes, demonstrating intra-class commonality and inter-class distinctiveness. As training and testing sets are randomly selected from these samples, this qualitative result proves that the proposed semantic regularization effectively fine-tunes the embedding network and better deal with covariate shift issues.

We also compare the similarity of training sample features against query sample features. This quantitatively measures the efficiency of the embedding networks which are fine-tuned by each baseline. In this work, we apply Fr\'{e}chet Inception Distance (FID) score~\cite{heusel2017gans_FID}, which is widely used in generative model evaluation, to measure the feature distribution difference between the training samples and testing samples. Ideally, with a close distribution, the classifier can perform better. The curves in Fig.~\ref{fig:test}-b show the result of 500 tests on covariate shift. For each test, 25 samples (5-way 5-shot) for the training set and 75 samples (15 queries/class) for the query set are randomly chosen to obtain its sample features by both methods. As shown in the curve figure, the embedding network fine-tuned by the proposed semantic regularization method consistently less than the embedding network fine-tuned by ProtoNet, indicating higher similarity, less covariate shift and a better embedding network.

\noindent\textbf{Metric module.} In addition, we further perform an ablation study to show the effectiveness of the semantic regularization module by simply applying Euclidean distance and relation network as metric (same as ProtoNet$^+$ and RelationNet$^+$).  The result is shown in Table~\ref{tab:proto_res_adn}. For clarity, we show the performance of all models with the same embedding network. It can be seen that due to the effectiveness of the proposed semantic regularization method, the network without relation module still achieves a competitive result against methods listed in Table~\ref{tab:Mini_result} and Table~\ref{tab:CUB_result}. Relation module combined with Euclidean distance further improves the performance.

The qualitative and quantitative results show that the proposed method can not only tackle the meta shift issue but also efficiently fine-tune the embedding network to alleviate the covariate shift issue in few-shot learning. It is also not limited by the metric choices. These benefits make the proposed method significantly outperform the baseline model and achieve competitive result against state-of-the-art methods.

\begin{table}[t]
\centering
\resizebox{0.9\columnwidth}{!}{
\begin{tabular}{c|l|c c }
\Xhline{3\arrayrulewidth}
\textbf{Dataset} &\textbf{Framework}          & \textbf{1-shot}     & \textbf{5-shot} \\
\Xhline{2\arrayrulewidth}
\multirow{4}{*}{MiniImageNet} & ResNet12+Euc     & 56.50 $\pm$ 0.40\% & 74.2 $\pm$ 0.20\%  \\
                              & ResNet12+ R    & 58.20 $\pm$ 0.30\% & 74.35 $\pm$ 0.23\%  \\
                              & ResNet12+SR+Euc      & 62.97 $\pm$ 0.16\% & 78.87 $\pm$ 0.15\%  \\
                              & ResNet12+SR+(Euc+R)  & \textbf{63.37 $\pm$ 0.14\%} & \textbf{79.03 $\pm$ 0.16\%}  \\
\Xhline{3\arrayrulewidth}
\multirow{4}{*}{CUB} & 4 Conv+Euc & 63.52 $\pm$ 0.25\% & 79.06 $\pm$ 0.20\% \\
                     &  4 Conv+R & 62.45 $\pm$ 0.98\% & 76.11 $\pm$ 0.69\% \\
                     & 4 Conv+SR+Euc & 66.80 $\pm$ 0.22\% & 82.0 $\pm$ 0.15\% \\
                     & 4 Conv+SR+(Euc+R) & \textbf{67.85 $\pm$ 0.24\%} & \textbf{83.53 $\pm$ 0.16\%} \\
\Xhline{3\arrayrulewidth}
\end{tabular}
}
\caption{Ablation study on two datasets by comparing ProtoNet$^+$ (Row1), RelationNet$^+$(Row2), our method without relation module (Row3), and with relation module (Row4).}
\label{tab:proto_res_adn}
\end{table}

\section{Conclusion and Future Work}
\label{sec:conclusion}

%
In this paper, to tackle the meta shift problem, we propose a concise Semantic Regularization Network under the meta-learning framework.
The proposed network is able to learn stable class descriptors from all support sets over the entire training process.
This is based on a novel encoder-decoder structure, which encodes the embedded samples into a semantic domain with trained semantic basis, and decodes back to the class domain.
We also utilize a metric module combining a fixed distance metric and a trainable relation module to classify the query sample according to the general class descriptors.
We evaluate the proposed framework by comparing with existing few-shot learning methods on three benchmark datasets. The state-of-the-art performance demonstrates the advantage of our work.

Several directions can be explored in the future. One way to improve the effectiveness of the semantic regularization module would be considering the relationship of the semantic basis. Another direction is to further strengthen the embedding network by using FPN~\cite{lin2017_fpn} or attention based techniques~\cite{wang2017residual}.


\bibliographystyle{plain}
\bibliography{egbib}
\end{document}